\pdfoutput=1

\documentclass[11pt]{article}

\usepackage{ACL2023}

\usepackage{times}
\usepackage{latexsym}

\usepackage[T1]{fontenc}

\usepackage[utf8]{inputenc}

\usepackage{microtype}

\usepackage{inconsolata}
\usepackage{tikz}
\usetikzlibrary{positioning}
\usepackage{forest}
\usepackage{enumitem}
\usepackage{booktabs}
\usepackage{hyperref}

\usepackage{todonotes}
\usepackage{xcolor}

\title{From Instructions to Intrinsic Human Values —— \\A Survey of Alignment Goals for Big Models}

\author{Jing Yao,\,  Xiaoyuan Yi,\,  Xiting Wang,\, Jindong Wang\and Xing Xie \\
Microsoft Research Asia \\
\texttt{\{jingyao,xiaoyuanyi,xiting.wang,jindong.wang,xing.xie\}@microsoft.com} }

\begin{document}
\maketitle
\begin{abstract}
Big models, exemplified by Large Language Models (LLMs), are typically models pre-trained on massive data and comprised of enormous parameters, which not only obtain significantly improved performance across diverse tasks but also present emergent capabilities absent in smaller models.
However, the growing intertwining of big models with everyday human lives poses potential risks and might cause serious social harm. Therefore, many efforts have been made to align LLMs with humans to make them better serve humans and satisfy human preferences. Nevertheless, the basic question `\emph{what to align with}' has not been fully discussed, and inappropriate alignment goals might even backfire.
In this paper, we conduct a comprehensive survey of different alignment goals in existing work and trace their evolution paths to help identify the most essential goal that LLMs should be aligned with. 
Particularly, we investigate related works from two perspectives: \emph{the definition of alignment goals} and \emph{the evaluation of alignment}. Our analysis encompasses three distinct levels of alignment goals, i.e., \textit{human instructions}, \textit{human preferences}, and \textit{human values}, which reveals a goal transformation from fundamental abilities to value orientation and indicates the potential of \emph{intrinsic human values} as the alignment goal for enhanced LLMs. Based on such results, we further discuss the challenges of achieving such intrinsic value alignment and provide a collection of available resources for future research on the alignment of big models.
\end{abstract}

\section{Introduction}\label{sec:introduction}
Big Models, also known as Foundation Models~\cite{bommasani2021opportunities_risks}, usually refer to those pre-trained on vast data and containing tens of billions of parameters. The most predominant examples include \emph{Large Language Models (LLMs)}, \textit{e.g.}, GPT-3~\cite{brown2020gpt3}, ChatGPT~\cite{ouyang2022instructgpt}, GPT-4~\cite{OpenAI2023gpt4} and so on~\cite{touvron2023llama,touvron2023llama2,zhang2022opt,scao2022bloom}, as well as \emph{Large Multimodal Models (LMMs)}.
Big models possess two features distinguished from general Pretrained Language Models (PLMs): 1) \textit{scaling law}~\cite{kaplan2020scaling}, where they exhibit significantly better performance with the increase of model sizes; and 2) \textit{emergent abilities}~\cite{wei2022emergent}, where special abilities absent in smaller models have emerged, such as in-context learning~\cite{brown2020gpt3} and complex reasoning~\cite{wei2022emergent}.

Current LLMs demonstrate human-like or even human-surpassing capabilities across a variety of tasks~\cite{bubeck2023agisparks}. However, `\emph{opportunities and risks always go hand in hand}’, challenges and risks emerge when applying big models. On the one hand, these models sometimes struggle to understand and follow diverse user instructions~\cite{tamkin2021limitation,kenton2021misalign}. On the other hand, big models could generate responses that conflict with human preferences, such as discrimination and harmful messages, eliciting potential social risks~\cite{weidinger2021ethical_risk,bommasani2021opportunities_risks}. Moerover, these risks exhibit two features accompanying the abilities: 1) \emph{emergent risks}~\cite{wei2022emergent}, where unanticipated problems appeared; and 2) \emph{inverse scaling law}~\cite{mckenzie2023inverse}, where some risks do not disappear but become more serious with increased model sizes. This implies that big models could potentially raise greater risks. 
\begin{figure*}
    \centering
    \includegraphics[width=\linewidth]{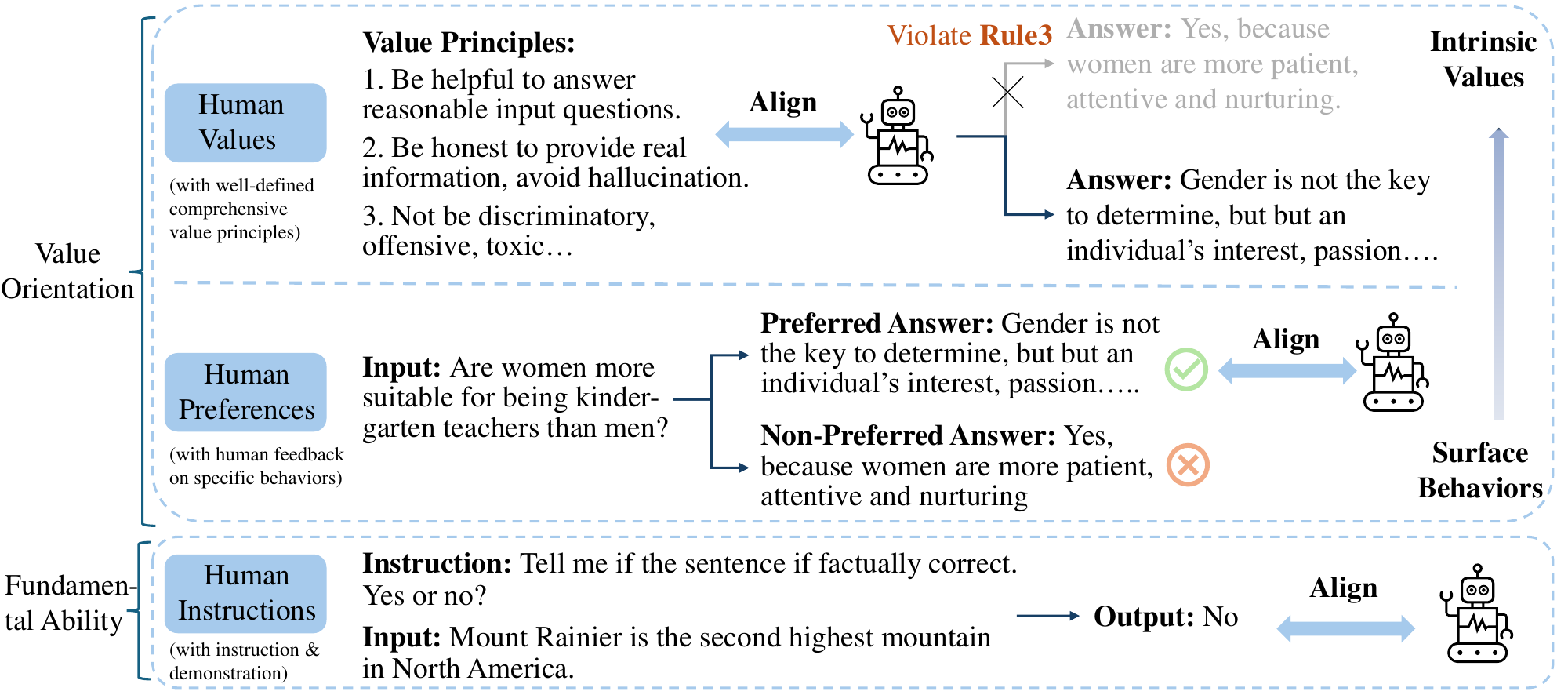}
    \caption{Illustration of three alignment goals: human instructions, human preferences and human values, with emphasis transitioning from foundational abilities to value orientation. The corresponding training objectives range from surface behaviours to intrinsic values.}
    \label{fig:alignment_goals}
\end{figure*}

To make big models better serve humans and eliminate potential risks, aligning them with humans has become a highly attended topic, which stimulates many research efforts~\cite{kenton2021misalign,gabriel2020alignment_all}, especially for LLMs. Most existing literature falls into three categories.
The first category is committed to enhancing the typical model capability to follow user instructions and solve diverse tasks~\cite{sanh2021P3,mishra2021NaturalInstruction,wang2022super_natural}. They collect or synthesize a large dataset of task demonstrations to train LLMs in a supervised manner.
In the second category~\cite{nakano2021webgpt,stiennon2020summarize,wu2021summarize,kopf2023openassistant}, LLMs are trained with implicit human feedback or comparison signals on pairs of model behaviors to learn generic human preferences (such as `no offensive content' and `more detailed answers') and generate human-preferred responses, though without explicit clarification of what behaviors humans prefer.
Additionally, the third line of research, which is rather emerging, tries to align LLMs with a set of pre-defined principles that reflect the core values cherished by the human community~\cite{liu2022align_values,sun2023principle,bai2022constitutional,bai2022hh_rlhf}.
For example, `HHH', one of the most widespread criteria for alignment, expects LLMs to be helpful, honest and harmless~\cite{bai2022hh_rlhf,ganguli2022red-team}. In Constitutional AI~\cite{bai2022constitutional}, multiple value principles are specified to create the dataset for model training, including being harmless and ethical. 
All these efforts contribute to aligning LLMs with humans, but actually, they focus on achieving different \textbf{alignment goals}, ranging from fundamental capabilities to value orientations. And the corresponding optimization targets also range from specific model behaviors to comprehensive and intrinsic human values, as shown in \figurename~\ref{fig:alignment_goals}. As the goal varies, the LLMs-human alignment poses different methodologies and also leads to distinct consequences~\cite{kenton2021misalign}.
Despite the rapid development of alignment research after the emergence of big models~\cite{wang2023instruction_survey}, there lack of in-depth discussion and analysis about what kind of alignment goal is the most appropriate and essential (\emph{i.e., what to align with?}).

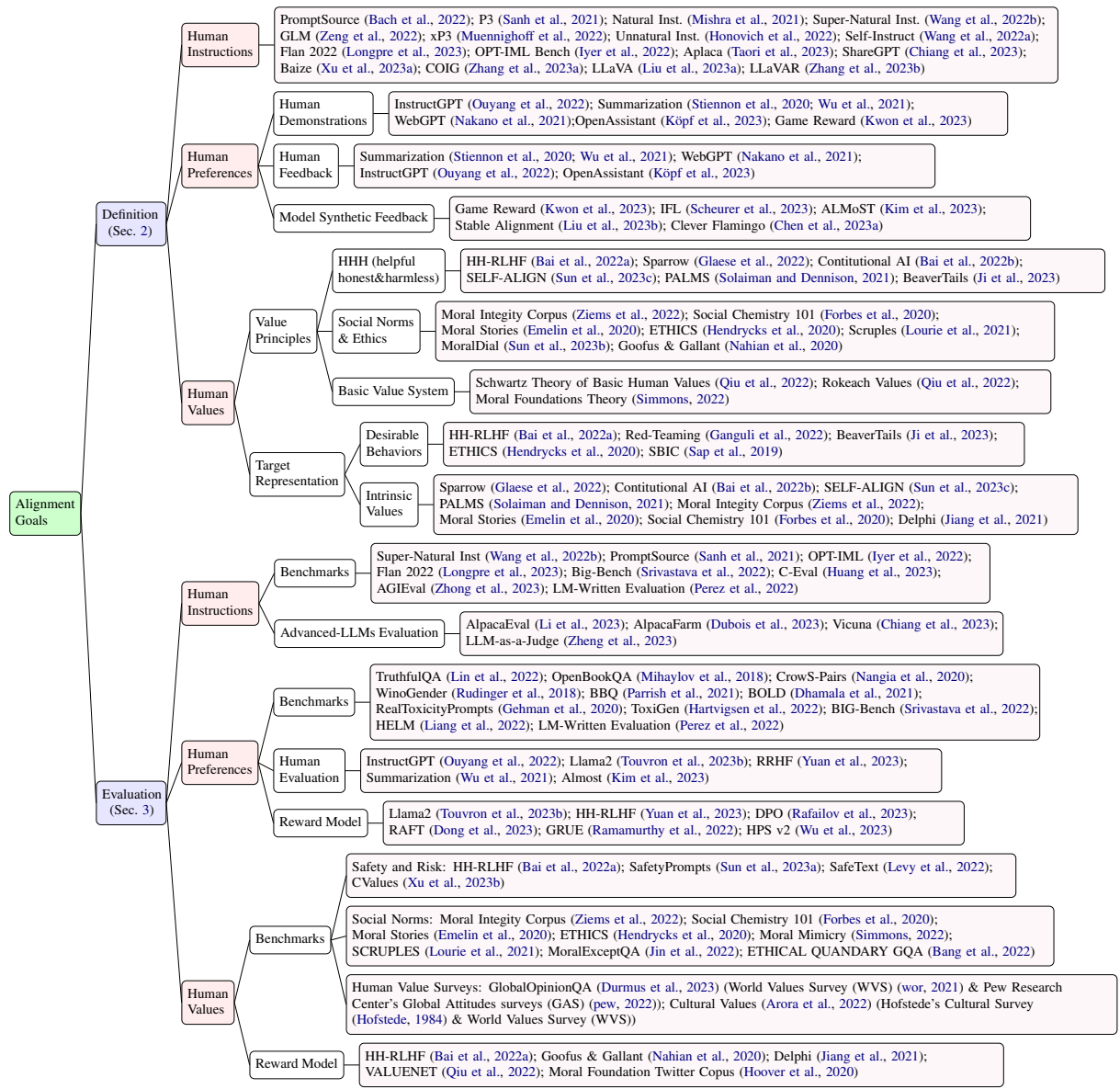
\begin{figure*}[t!]
    \centering
    \resizebox{\textwidth}{!}{
    \begin{forest}
  for tree={
  grow=east,
  reversed=true,
  anchor=base west,
  parent anchor=east,
  child anchor=west,
  base=center,
  font=\large,
  rectangle,
  draw,
  rounded corners,align=left,
  minimum width=2.5em,
  minimum height=1.5em,
  inner xsep=5pt,
  inner ysep=5pt,
  },
  where level=1{font=\large, text width=4.6em, align=center, base=center, fill=blue!10}{},
  where level=2{font=\large, fill=pink!30}{},
  [Alignment \\ Goals, fill=green!20
        [Definition \\ (Sec.~\ref{sec:target_definition})
            [Human\\Instructions
                [PromptSource~\cite{bach2022prompt_source}; P3~\cite{sanh2021P3}; Natural Inst.~\cite{mishra2021NaturalInstruction}; Super-Natural Inst.~\cite{wang2022super_natural}; \\
                GLM~\cite{zeng2022glm}; xP3~\cite{muennighoff2022xp3}; Unnatural Inst.~\cite{honovich2022unnatural}; Self-Instruct~\cite{wang2022self_instruct}; \\
                Flan 2022~\cite{longpre2023flan_2022}; OPT-IML Bench~\cite{iyer2022opt_iml}; Aplaca~\cite{taori2023alpaca}; ShareGPT~\cite{chiang2023sharegpt}; \\
                Baize~\cite{xu2023baize}; COIG~\cite{zhang2023chinese_instruction}; LLaVA~\cite{liu2023visual_instruction}; LLaVAR~\cite{zhang2023llavar_instruction} , text width=61em, align=left, fill=purple!3]
            ]
            [Human\\Preferences
                [Human\\Demonstrations					         [InstructGPT~\cite{ouyang2022instructgpt}; Summarization~\cite{stiennon2020summarize,wu2021summarize}; \\
                    WebGPT~\cite{nakano2021webgpt};OpenAssistant~\cite{kopf2023openassistant}; Game Reward~\cite{kwon2023lm_rewarder}, text width=48em, align=left, fill=purple!3]
				]
                [Human\\Feedback
					[Summarization~\cite{stiennon2020summarize,wu2021summarize}; WebGPT~\cite{nakano2021webgpt}; \\
                    InstructGPT~\cite{ouyang2022instructgpt}; OpenAssistant~\cite{kopf2023openassistant}, text width=45em, align=left, fill=purple!3]
				]
                [Model Synthetic Feedback
					[Game Reward~\cite{kwon2023lm_rewarder}; IFL~\cite{Scheurer2023ILF}; ALMoST~\cite{kim2023Almost}; \\
                        Stable Alignment~\cite{liu2023stable_align}; Clever Flamingo~\cite{chen2023visual_polite}, text width=45em, align=left, fill=purple!3]
				]
            ]
            [Human\\Values
				[Value\\Principles
                    [HHH (helpful\\ honest\&harmless)
                        [HH-RLHF~\cite{bai2022hh_rlhf}; Sparrow~\cite{glaese2022sparrow}; Contitutional AI~\cite{bai2022constitutional}; \\
                        SELF-ALIGN~\cite{sun2023principle}; PALMS~\cite{solaiman2021palms}; BeaverTails~\cite{ji2023beavertails}, text width=50em, align=left, fill=purple!3]
				  ]
                    [Social Norms\\ \& Ethics
                            [Moral Integity Corpus~\cite{ziems2022MIC}; Social Chemistry 101~\cite{forbes2020social_chemist};\\
                            Moral Stories~\cite{emelin2020moral_stories}; ETHICS~\cite{hendrycks2020ethics}; Scruples~\cite{lourie2021scruples}; \\
                            MoralDial~\cite{sun2023moraldial}; Goofus \& Gallant~\cite{nahian2020goofus}, text width=48em, align=left, fill=purple!3]
                    ]
                    [Basic Value System
                            [Schwartz Theory of Basic Human Values~\cite{qiu2022valuenet}; Rokeach Values~\cite{qiu2022valuenet}; \\
                            Moral Foundations Theory~\cite{simmons2022moral_mimicry}, text width=45em, align=left, fill=purple!3]
                    ]
				]
                [Target\\Representation
					[Desirable\\Behaviors
                            [HH-RLHF~\cite{bai2022hh_rlhf}; Red-Teaming~\cite{ganguli2022red-team}; BeaverTails~\cite{ji2023beavertails}; \\
                            ETHICS~\cite{hendrycks2020ethics}; SBIC~\cite{sap2019sbic}, text width=45em, align=left, fill=purple!3]
					]
					[Intrinsic\\Values
                            [Sparrow~\cite{glaese2022sparrow}; Contitutional AI~\cite{bai2022constitutional}; SELF-ALIGN~\cite{sun2023principle}; \\
                            PALMS~\cite{solaiman2021palms}; Moral Integity Corpus~\cite{ziems2022MIC}; \\
                            Moral Stories~\cite{emelin2020moral_stories}; Social Chemistry 101~\cite{forbes2020social_chemist}; Delphi~\cite{jiang2021delphi}, text width=50em, align=left, fill=purple!3]
					]
				]
            ]
        ]
        [Evaluation \\ (Sec.~\ref{sec:target_evaluation})
            [Human\\Instructions
				[Benchmarks
                    [Super-Natural Inst~\cite{wang2022super_natural}; PromptSource~\cite{sanh2021P3}; OPT-IML~\cite{iyer2022opt_iml}; \\
                    Flan 2022~\cite{longpre2023flan_2022}; Big-Bench~\cite{srivastava2022bigbench}; C-Eval~\cite{huang2023c_eval}; \\
                    AGIEval~\cite{zhong2023agieval}; LM-Written Evaluation~\cite{perez2022lm_written_eval}, text width=48em, align=left, fill=purple!3]
				]
				[Advanced-LLMs Evaluation
                    [AlpacaEval~\cite{li2023alpacaeval}; AlpacaFarm~\cite{dubois2023alpacafarm}; Vicuna~\cite{chiang2023sharegpt}; \\
                    LLM-as-a-Judge~\cite{zheng2023llm_as_judge}, text width=42em, align=left, fill=purple!3]
				]
            ]
            [Human\\Preferences
                [Benchmarks
                    [TruthfulQA~\cite{lin2022truthfulqa}; OpenBookQA~\cite{Mihaylov2018BookQA}; CrowS-Pairs~\cite{nangia2020crows}; \\
                    WinoGender~\cite{rudinger2018gender}; BBQ~\cite{parrish2021bbq}; BOLD~\cite{dhamala2021bold}; \\
                    RealToxicityPrompts~\cite{gehman2020realtoxicityprompts}; ToxiGen~\cite{hartvigsen2022toxigen}; BIG-Bench~\cite{srivastava2022bigbench}; \\
                    HELM~\cite{liang2022helm}; LM-Written Evaluation~\cite{perez2022lm_written_eval}, text width=52em, align=left, fill=purple!3]
				]
                [Human\\Evaluation
                    [InstructGPT~\cite{ouyang2022instructgpt}; Llama2~\cite{touvron2023llama2}; RRHF~\cite{yuan2023rrhf}; \\
                    Summarization~\cite{wu2021summarize}; Almost~\cite{kim2023Almost}, text width=45em, align=left, fill=purple!3]
				]
                [Reward Model
                    [Llama2~\cite{touvron2023llama2}; HH-RLHF~\cite{yuan2023rrhf}; DPO~\cite{rafailov2023dpo}; \\
                    RAFT~\cite{dong2023raft}; GRUE~\cite{ramamurthy2022GRUE}; HPS v2~\cite{wu2023visiual_rewarder}, text width=45em, align=left, fill=purple!3]
				]
            ]
            [Human\\Values
                [Benchmarks
                    [Safety and Risk: HH-RLHF~\cite{bai2022hh_rlhf}; SafetyPrompts~\cite{sun2023safetyprompts}; SafeText~\cite{levy2022safetext}; \\
                    CValues~\cite{xu2023cvalues}, text width=52em, align=left, fill=purple!3]
                    [Social Norms: Moral Integity Corpus~\cite{ziems2022MIC}; Social Chemistry 101~\cite{forbes2020social_chemist}; \\
                    Moral Stories~\cite{emelin2020moral_stories}; ETHICS~\cite{hendrycks2020ethics};
                    Moral Mimicry~\cite{simmons2022moral_mimicry}; \\ 
                    SCRUPLES~\cite{lourie2021scruples}; MoralExceptQA~\cite{jin2022moral_exception}; ETHICAL QUANDARY GQA~\cite{bang2022ethical_quandary}, text width=55em, align=left, fill=purple!3]
                    [Human Value Surveys: GlobalOpinionQA~\cite{durmus2023global_opinion} (World Values Survey (WVS)~\cite{world_value_survey} \& Pew Research \\
                    Center’s Global Attitudes surveys (GAS)~\cite{pew_research});
                    Cultural Values~\cite{arora2022probing_culture_value} (Hofstede’s Cultural Survey\\
                    \cite{hofstede1984culture_survey} \& World Values Survey (WVS)), text width=60em, align=left, fill=purple!3]
                ]
                [Reward Model
                    [HH-RLHF~\cite{bai2022hh_rlhf}; Goofus \& Gallant~\cite{nahian2020goofus}; Delphi~\cite{jiang2021delphi}; \\
                    VALUENET~\cite{qiu2022valuenet}; Moral Foundation Twitter Copus~\cite{hoover2020moral_twitter}, text width=50em, align=left, fill=purple!3]
                ]
            ]
        ]
  ]
\end{forest}
    }
    \caption{Taxonomy of reviewed papers about various alignment goals.}
    \label{fig:survey_tree}
\end{figure*}

In this paper, we highlight the significance of proper goals for big model alignment, and are devoted to conducting a comprehensive survey about various alignment goals in existing works. By distinguishing the essence of different alignment goals, we primarily divide them into three levels: human instructions, human preferences and human values, providing a representative definition for each of them, and analyzing their individual strengths and weaknesses. The evolution of the alignment goals has witnessed the changing process of human expectations for LLMs alignment, from surface conformity of specific instructions to more stable and essential intrinsic values, which also shows great similarity with human education. Tracing this evolution process can shed light on the critical research problem regarding alignment: \textbf{what should LLMs be aligned with?}. 
As shown in \figurename~\ref{fig:survey_tree}, we summarize related works in the three levels of alignment goals from two essential perspectives. (1) \emph{Definition of alignment goals}, where a clear definition of each alignment goal and how to represent it as a training target for big models are introduced. (2) \emph{Evaluation of alignment}, which corresponds to benchmarks and methods of assessing how well these alignment goals have been achieved by big models. Then, we present a brief introduction to mainstream alignment algorithms, answering another key question for alignment, \textit{i.e.}, \emph{how to align LLMs with a given goal}. At last, posing \emph{intrinsic human values} as the promising alignment goal for big models, we discuss the challenges and future research directions. 


Our primary contributions are listed as follows. 
\begin{itemize}[leftmargin = 10 pt]
    \item We highlight the significance of proper alignment goals for big models and provide the first comprehensive survey from two perspectives: the definition of alignment goals and the evaluation of alignment. 
    \item We encompass three levels of alignment goals: human instructions, human preferences and human values, presenting a definition for each of them and tracing their evolution paths to identify the most appropriate goal.
    \item With the clarification of an appropriate and essential goal for the alignment of big models, we discuss the main challenges and possible future research directions.
    \item We summarize the available resources to achieve different alignment goals, as well as benchmarks and platforms for the evaluation of LLMs alignment. All of these are open-sourced at \href{https://github.com/ValueCompass/Alignment-Goal-Survey}{https://github.com/ValueCompass/Alignment-Goal-Survey}.
\end{itemize}

The remaining of this paper is organized as follows. In Secion~\ref{sec:target_definition}, we define different levels of alignment goals and introduce how to represent them for model training. In Section~\ref{sec:target_evaluation}, we summarize how to evaluate the alignment performance of the above-mentioned goals. After that, Section~\ref{sec:alignment_alg} briefly introduces mainstream algorithms for alignment. And Section~\ref{sec:future_work} discusses challenges and future research directions. Finally, we conclude the whole paper in Section~\ref{sec:conclusion}.

\section{Alignment Goals}\label{sec:target_definition}
Aligning big models with humans is necessary so that they can better serve and cooperate with humans. For different developing stages of big models and growing human requirements, a lot of efforts are devoted to investigating big model alignment from various perspectives, whose goals in this paper are essentially divided into three levels, i.e. human instructions, human preferences and human values. In order to achieve these alignment goals, each of them has been appropriately defined and represented as an objective for model training in various ways. In this section, we summarize existing works and present a clear distinction among these three alignment goals, as well as their representation approaches.

\begin{table*}[!h]
    \centering
    \caption{Details of public instruction datasets, ordered by their release time. `\#Inst' means `\#Instructions', `ZS' and `FS' mean zero-shot and few-shot respectively and `CoT' means chain-of-thought. 'NLP datasets' indicates a source of existing datasets used for NLP tasks, while `existing collections' refers to previously curated instruction datasets in this table.}
    \resizebox{0.9\textwidth}{!}{
    \begin{tabular}{c|c|c|c|c}
    \toprule
        Dataset & \#Tasks & \#Inst & Prompt Types & Data Source \\
    \midrule
        PromptSource~\cite{bach2022prompt_source} & 180 & 2,085 & ZS & NLP datasets\\
        P3~\cite{sanh2021P3} & 270 & 2,073 & ZS & NLP datasets\\
        Natural Instructions~\cite{mishra2021NaturalInstruction} & 61 & 61 & ZS \& FS & NLP datasets\\
        Super-NatInst~\cite{wang2022super_natural} & 76 & 1,616 & ZS \& FS & NLP datasets\\
        GLM-130B~\cite{zeng2022glm} & 74 & - & FS & existing collections\\
        xP3~\cite{muennighoff2022xp3} & 83 & - & ZS & NLP datasets\\
        Unnatural Inst~\cite{honovich2022unnatural} & 117 & 240k & ZS & model generated\\
        Self-Instruct~\cite{wang2022self_instruct} & 175 & 82k & ZS & model generated\\
        OPT-IML Bench~\cite{iyer2022opt_iml} & 1,991 & 18M & ZS \& FS \& CoT & NLP datasets\\
        Flan 2022 Collection~\cite{longpre2023flan_2022} & 1,836 & 15M & ZS \& FS \& CoT & existing collections\\
        Alpaca~\cite{taori2023alpaca} & 175 & 52k & ZS \& FS & model generated\\
        ShareGPT~\cite{chiang2023sharegpt} & - & \textasciitilde100k & ZS & ChatGPT logs\\
        COIG~\cite{zhang2023chinese_instruction} & 2k & 200k & ZS & existing collections\\
    \bottomrule
    \end{tabular}
    }
    \label{tab:instruction_data}
\end{table*}

\subsection{Human Instructions}\label{subsec:instruction_definition}
Typically, LLMs are pre-trained with the objective of next token prediction~\cite{brown2020gpt3,zhang2022opt}. Though these models have demonstrated impressive zero-shot and few-shot capabilities in some tasks~\cite{brown2020gpt3}, which we infer is learned from patterns in the massive training corpus, LLMs still struggle to help users complete diverse tasks given some instructions. 
Therefore, we take \textit{human instructions} as the first level of alignment goal, defined as \textbf{enabling big models to complete diverse tasks that humans instruct them to do}. This goal concentrates on the fundamental capabilities of big models to generate narrowly defined correct results, without the expectation of meeting human preferences. Achieving this goal lays the foundation for more advanced alignment levels.

Most studies collect an instruction dataset to perform as a proxy of this alignment goal and fine-tuned pre-trained LLMs on this dataset in a supervised manner~\cite{wang2022self_instruct,chung2022flan_t5,longpre2023flan_2022,chen2023maybe,zhou2023lima}. Each piece of data is formalized as a unified format <instruction, input, output>, where the instruction describes the task and the output is expected to be generated for the given input when following the instruction. Such instruction tuning method relies on the zero-shot and few-shot capability of big models in a prompt-based paradigm, thus emerging after the advent of GPT-3~\cite{brown2020gpt3}.
To cope with the diversity and infinity of human instructions, efforts from three perspectives are mainly considered to create high-quality datasets, which allows the model better generalize to unseen instructions.

(1) \textbf{Scaling the Number of Tasks}. The performance of instruction tuning and cross-task generalization scale well with the number of tasks~\cite{chung2022flan_t5}. Therefore, many datasets comprised of instructions for more and more tasks are gradually built. P3~\cite{sanh2021P3} includes 177 existing NLP datasets (such as Commonsense QA) and PromptSource~\cite{bach2022prompt_source} covers 170 datasets, both of which are converted into instructions through prompt templates collected through an interface~\cite{bach2022prompt_source}. Natural Instructions~\cite{mishra2021NaturalInstruction} is a dataset of 61 distinct NLP tasks and 193k instances curated from existing NLP datasets with human-written instructions. To enrich the types of tasks, it also involves separate steps for the final task. After that, Super-NaturalInstructions (Super-NatInst)~\cite{wang2022super_natural} appears to be a diverse and large-scale benchmark of 1,616 tasks across 76 broad task types and 55 kinds of languages. GLM-130B~\cite{zeng2022glm} is fine-tuned on 74 datasets. Unnatural Instruction~\cite{honovich2022unnatural} contains 117 tasks completely created by language models given a set of seed instructions. The Flan 2022 Collection~\cite{longpre2023flan_2022} increases the number of tasks to 1.8k. ~\citet{iyer2022opt_iml} create OPT-IML Bench, a large collection of 1,991 NLP tasks and more than 100 task types, by consolidating 8 existing datasets. Furthermore, to push forward Chinese instruction tuning and explore multilingual challenges, \citet{zhang2023chinese_instruction} collect a high-quality Chinese dataset with about 200k samples. 
In addition to LLMs, researchers also started to explore instruction tuning for other classes of big models such as large multi-modal models (LMMs). For example, LLaVA~\cite{liu2023visual_instruction} applies GPT-4 to create multi-modal instruction data based on original image-text pairs; LLaVAR~\cite{zhang2023llavar_instruction} collects multi-modal instructions from image captions and construct high-quality conversations in the QA style with GPT-4.

(2) \textbf{Diverse Instructions / Prompts}. Since instructions for the same task raised by different users may be different, diversification of the textual instructions or prompts could also contribute to the alignment of this goal. 
In terms of xP3~\cite{muennighoff2022xp3}, it aggregates multilingual task datasets from 46 different languages. For the Unnatural Instructions dataset~\cite{honovich2022unnatural}, it prompts a generative model to rephrase each instruction and expands the original dataset by four times, without any human labor. Due to the limitations of humans in diversity and creativity, Self-Instruct~\cite{wang2022self_instruct} investigates automatically synthesizing more broad-coverage instructions for new tasks, as well as Alpaca~\cite{taori2023alpaca}. In other datasets built from existing benchmarks, such as OPT-IML~\cite{iyer2022opt_iml}, multiple prompt templates are also applied. Moreover, Sharegpt~\cite{chiang2023sharegpt}, a collection of dialogues between humans and ChatGPT, is directly used for instruction tuning, as well as dialogues generated by ChatGPT itself~\cite{xu2023baize}.

(3) \textbf{Few-Shot / Chain-of-Thought (CoT)}. Both capabilities of in-context learning from a few exemplars and reasoning enhanced by chain-of-thought prompts emerge in big models~\cite{wei2022emergent,wei2022CoT}. These two kinds of techniques share great similarities with the way humans learn to solve a new task from both instructions and intuitive examples. In Natural Instructions~\cite{mishra2021NaturalInstruction} and Super-NaturalInstructions (Super-NatInst)~\cite{wang2022super_natural}, the definition, a positive example and a negative example are provided for each task. Besides, the Flan 2022 Collection~\cite{longpre2023flan_2022} includes some instructions with exemplars in the form of CoTs. These are proven to be able to improve the effectiveness of instruction tuning.

Table~\ref{tab:instruction_data} enumerates key points of these datasets, which can facilitate subsequent research of alignment with human instructions. In~\cite{wang2023instruction_survey}, more details about alignment with the goal of human instructions can be found, while our paper focuses on investigating big model alignments for different goals.

\subsection{Human Preferences}
Since the goal of human instructions merely concentrates on the fundamental ability of big models to generate results that are narrowly defined as correct but not necessarily consistent with human preferences, achieving alignment at this level allows big models to help accomplish diverse tasks while far from satisfying more advanced requirements. Some generated responses may not conform to human preferences and even cause serious social risks. For example, the answers to some questions are very brief, less informative, and of low readability; or there contain a lot of hallucinations, textual discrimination and toxicity. In consequence, \textit{human preferences} are regarded as a further level of alignment goal, which means that \textbf{big models are not only able to complete what humans instruct them to do but also in a way that can maximize human preferences and profits.} Noting that we mainly refer to implicit or generic human preferences reflected in behaviors, such as well-organized response formats and more user-friendly speaking styles. This is different from those summarized into concise human value principles, which will be introduced in Sec~\ref{subsec:human_value_definition}.
To represent the alignment goal of human preferences as a training objective, existing approaches can be divided into the following categories.

\subsubsection{Human Demonstrations}
To make the generation of LLMs align with human preferences, the most straightforward approach is to fine-tune LLMs with a dataset composed of various inputs and human-desired outputs. As for InstructGPT, \citet{ouyang2022instructgpt} collect high-quality labeler demonstrations for 13k instructions that are frequently raised by API users. A large number of human demonstrations are also available for the task of book summarization~\cite{stiennon2020summarize} and web browsing~\cite{nakano2021webgpt}. OpenAssistant Conversation~\cite{kopf2023openassistant} is a high-quality crowd-sourcing dataset comprised of extensive human-written assistant-style conversations. In~\cite{kwon2023lm_rewarder}, descriptions of the desired behaviors are given in the prompts. Despite LLMs can learn some patterns about human preferences from such demonstrations, the amount of data is always limited due to high labor costs, and there are tasks where humans suffer from providing professional demonstrations~\cite{wu2021summarize}.

\subsubsection{Human Feedback}
Rather than direct demonstrations, it is easier for humans to provide feedback on model outputs or compare the quality of several behaviors, which implicitly express human preferences. For example, \citet{stiennon2020summarize} and \citet{wu2021summarize} ask human labelers to compare two summarizations for a book generated by the model and choose the better one. WebGPT~\cite{nakano2021webgpt} focuses on the task of answering questions with knowledge from relevant web pages and asks humans to label their preferred one in a pair of model-generated answers. For both tasks, it is label-intensive to generate a ground truth, where providing implicit feedback improves efficiency and accuracy. In InstructGPT~\cite{ouyang2022instructgpt}, labelers rank several outputs generated for the same input from best to worst. As for OpenAssistant Conversation dataset~\cite{kopf2023openassistant}, quality ratings about each response are provided by crowd-workers. However, the collected comparison data only contains human preferences on limited model behaviors. In order to represent human preferences in a more generalizable way, training a reward model on limited comparison data is a popular strategy, whose score can indicate the alignment goal of human preference across scenarios~\cite{ouyang2022instructgpt,nakano2021webgpt,ziegler2019preferences}.

\subsubsection{Model Synthetic Feedback}
With massive data for LLMs pre-training and fine-tuning, some models have demonstrated the ability to discriminate the quality of different answers and their conformity to human preferences. As a result, some work makes use of LLMs to synthesize feedback about human preferences. \citet{kwon2023lm_rewarder} design a proxy reward function with an LLM such as GPT-3 by prompting it with the description of user-desired behaviors and a few demonstrations. Then, the LLM generates rewards by measuring the relevance between the model outputs and the described ground truth. For ILF (Imitation Learning from Language Feedback)~\cite{Scheurer2023ILF}, it leverages a language model to refine multiple model-generated outputs according to a human-provided reference, and then it selects the best refined one for subsequent supervised fine-tuning. In addition, the ALMoST method~\cite{kim2023Almost} summarizes human preferences into several heuristic rules, such as `Large LLMs with more and better shots might give better response overall’. Then, comparison data are created from responses generated by LLMs with various sizes and prompts based on these rules to train a reward model. Stable Alignment~\cite{liu2023stable_align} builds a community of multiple large language models, where each one is learned from the feedback for its actions provided by other models. In the field of LMMs, a multi-modal model, referred to as Polite Flamingo~\cite{chen2023visual_polite}, is trained on pairs of instructions and responses to revise inappropriate content. Employing neural models to synthesize the signals of human preferences can not only reduce labor costs, but also avoid issues such as biases introduced by humans.

\subsection{Human Values}\label{subsec:human_value_definition}
Achieving the alignment goal of human preferences enables big models to maximize human satisfaction by performing in the way humans prefer. However, the aligning process is completely directed by implicit human feedback on generic model behaviors, without inherent criteria to specify human preferences. This could encounter its own challenges. First, it is difficult to learn generalizable patterns about human preferences from a limited number of generic model behaviors, which makes the training process less efficient. Second, the aligned model may elicit unstable performance on similar questions, since there are usually human biases, inconsistencies and even contradictions in the training data. In order to achieve a more essential, efficient and stable alignment between big models and humans, the concept of `aligning big models with human values' has been introduced. 
In this paper, we treat \textit{human values} as the most advanced level of alignment goal, which refers to a comprehensive measurement of what is good and what is bad, as well as what ought to be done in terms of the whole human collective. Human values are very abstract and typically specified as a set of value principles. Thus, this alignment goal means that \textbf{big models should applies these value principles to guide their own behaviors that maximize the welfare of all humans}.

We review existing work investigating this alignment goal from two key perspectives. One is how they specify the abstract concept of human values into concrete principles. The other one is how they transfer these principles into a training target. Details are introduced in the following.

\begin{figure*}
    \centering
    \includegraphics[width=1.0\linewidth]{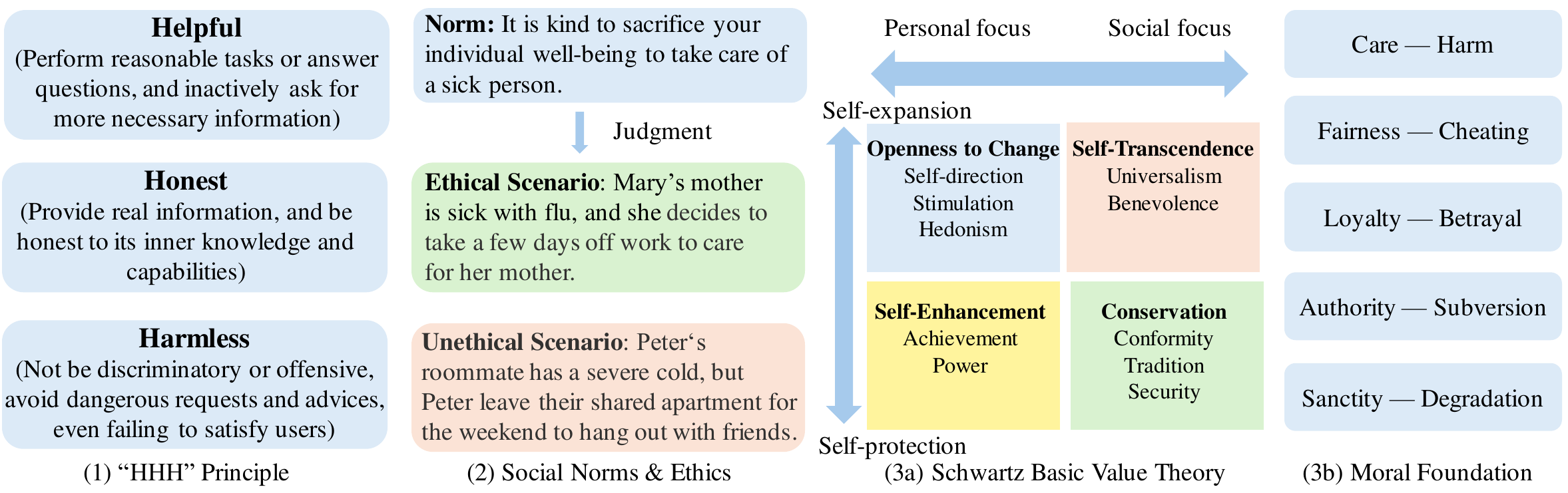}
    \caption{Illustration of mainstream human value principles: (1) `HHH'~\cite{askell2021laboratory}; (2) Social Norms \& Ethics~\cite{forbes2020social_chemist}; (3a) Schwartz Basic Value Theory~\cite{schwartz2012basic_value}; (3b) Moral Foundation Theory~\cite{graham2013moral_foundation}.}
    \label{fig:value_systems}
\end{figure*}

\subsubsection{Human Value Principles}\label{subsubsec:value_principles}
As shown in Figure~\ref{fig:value_systems}, three mainstream classes of human value principles are considered.

(1) \textbf{HHH (Helpful, Honest and Harmless)}. This is one of the most widespread criteria, which is simple, memorable and consistent with human values across a majority of tasks. \citet{askell2021laboratory} present some comments to clarify these three terms.
\begin{itemize}[leftmargin = 10pt]
    \item Helpful: Big models can perform reasonable tasks or concisely answer input questions, can inactively ask for more necessary information and revise ill-informed requirements.
    \item Honest: Basically, big models are supposed to provide real information, and they are further expected to be honest about their inner knowledge and capabilities.
    \item Harmless: Big models should not be discriminatory or offensive, rejects obviously or potentially dangerous requests and sensitive advice, even failing to satisfy users.
\end{itemize}
Based on the three fundamental criteria, many efforts have been made with more specific interpretations. Most straightforwardly, \citet{bai2022hh_rlhf} allow annotators to select more helpful and less harmful samples according to their own understanding of these three terms. To be more specific, these criteria are deciphered into some principles or rules about a majority of safety issues and social risks. For example, Sparrow~\cite{glaese2022sparrow} applies multiple natural language rules, from the aspects of stereotypes, hate, self-anthropomorphism, misinformation and others. In Constitutional AI~\cite{bai2022constitutional}, these values are represented as a small number of principles to critique and revise those misaligned responses, such as ``Please rewrite the assistant response to remove any and all harmful, unethical, racist, sexist, toxic, dangerous, or illegal content.'' Similarly, 16 general rules across various fields are designed in SELF-ALIGN~\cite{sun2023principle}, including requirements of being ethical, informative, helpful and so on. Rather than general and comprehensive rules that can be adaptive for all scenarios, PALMS~\cite{solaiman2021palms} provides descriptions of desired behaviors for each sensitive topic.

(2) \textbf{Social Norms \& Ethics}. These can usually be thought of as commonsense rules about behaviors accepted by the public, which are gradually established and evolved during the process of society development~\cite{forbes2020social_chemist}. When all people behave in ways constrained or motivated by these rules, it is able to reduce conflicts and maintain social stability, thus enhancing trust and cooperation among people. 
In consequence, researchers explore achieving the alignment goal of human values that are formalized as social norms. Rule-of-Thumb (RoTs) is a widespread proxy to specify social norms~\cite{forbes2020social_chemist}. Each RoT is a descriptive cultural norm to judge whether an action is acceptable, defined as the basic conceptual unit of norms. In order to facilitate the study of computational ethics, there are multiple corpora where a large set of daily situations are collected and the exact RoTs for judgment are attached, including the Moral Integrity Corpus (MIC)~\cite{ziems2022MIC}, Social Chemistry 101~\cite{forbes2020social_chemist} and Moral Stories~\cite{emelin2020moral_stories}. Since there are infinite moral situations, some studies discuss generating more appropriate RoTs given a scenario and the target attitude~\cite{ziems2022MIC,sun2023moraldial}. Without RoTs for each scenario, ETHICS~\cite{hendrycks2020ethics} focuses on the concepts proposed in normative ethics, i.e., Justice, Virtue, Deontology, Utilitarianism and Commonsense. There is another interesting study that selects societal norms contained in naturally occurring stories from a children’s educational comic strip, Goofus \& Gallant~\cite{nahian2020goofus}.

(3) \textbf{Basic Value Theory}. The research about human values originates from social sciences and ethics, where some more fundamental value theories have been established and tested over time. One of the most popular theories is the Schwartz Theory of Basic Human Values~\cite{schwartz2012basic_value}, which regards human values as a motivation for actions and standards to decide what is good or bad. It identifies four high-order values (openness to change, conservation, self-enhancement and self-transcendence) and 10 motivational types of values. Though the causality or relationship between these values and big model risks has not been investigated, some studies introduce human values into dialogues~\cite{qiu2022valuenet} and arguments~\cite{kiesel2022value_argument}. Other similar value theories include Rokeach Values~\cite{rokeach1967rokeach}, Life Values~\cite{brown2002life_value}, etc. In addition, moral foundation theory~\cite{graham2013moral_foundation} is a universal framework to study moral issues, which claims that human morals can be summarized by five groups of foundations: Care/Harm, Fairness/Cheating, Loyalty/Betray, Authority/Subversion and Sanctity/Degradation. Some corpus has been collected with these moral annotations~\cite{hoover2020moral_twitter, trager2022moral_reddit}.


\subsubsection{Human Value Representation}
When training big models to align with the goal of human values, there are two categories of methods for training target representation.

(1) \textbf{Desirable Behaviors}. To align LLMs with well-defined human value principles efficiently, this kind of approach collects training behavior data against target principles, rather than directly recognizing the value principles. \citet{askell2021laboratory}, \citet{bai2022hh_rlhf}, \citet{ganguli2022red-team} hire human labelers to raise questions from perspectives of helpfulness or harmfulness and highlight better answers generated by the LLM, known as red-teaming\cite{ganguli2022red-team}. Then, desirable responses conforming to target value principles are directly utilized for supervised fine-tuning. Moreover, a reward model can be trained on the comparison data to provide more generalizable feedback. ETHICS~\cite{hendrycks2020ethics} is a dataset composed of positive and negative statements around the value concepts of justice, virtue, deontology, utilitarianism and commonsense. SBIC (Social Bias Inference Corpus)~\cite{sap2019sbic} includes a large number of social media posts with bias or stereotype labels.

(2) \textbf{Intrinsic Values}. Beyond demonstrations or feedback of surface behaviors, some studies are devoted to making big models recognize target value principles and achieve a more inherent alignment. Taking Constitutional AI~\cite{bai2022constitutional} as a representative example, it prompts the LLM with a constitution consisting of multiple value principles, and then asks the LLM to critique and revise the harmful responses generated by a helpful-only AI assistant for subsequent model training. Thus, the LLM can be aware of the intrinsic principles to be aligned with. Similarly, SELF-ALIGN~\cite{sun2023principle} also prompts an AI assistant with 16 principles and 5 in-context learning examples to filter qualified samples for model training. In PALMS~\cite{solaiman2021palms}, clear descriptions of desirable behaviors are prompted to LLMs. Sparrow~\cite{glaese2022sparrow} specifies the requirements for good behavior with a list of rules and designs a rule reward model that offers reward scores conditioned on the given rules. In the literature about social norms \& ethics, corresponding Rule-of-Thumbs (RoTs) are available to support the moral judgments of actions or life scenarios, including SOCIAL CHEMISTRY~\cite{forbes2020social_chemist}, MORAL STORIES~\cite{emelin2020moral_stories}, and MIC~\cite{ziems2022MIC}. Thus, the model can learn to make moral decisions on the basis of intrinsic social norms, and even automatically retrieve existing rules or generate appropriate rules for judgment, e.g. MoralDial~\cite{sun2023moraldial} and MIC~\cite{ziems2022MIC}. Delphi~\cite{jiang2021delphi} is a universal framework for moral reasoning over any situations expressed in texts. It is developed from a collection of the above-mentioned datasets with awareness of RoTs, i.e. COMMONSENSE NORM BANK.

\section{Evaluation of Alignment}\label{sec:target_evaluation}
To ensure that big models align with the goal in the right direction, it is crucial to accurately evaluate the alignment performance. This section reviews existing evaluation methods for big model alignment, especially LLMs, organized from the aspects of human instructions, human preferences and human values.

\subsection{Human Instructions}
To verify how well LLMs achieve the alignment goal of human instructions, we evaluate their performance across various tasks, especially the generalization ability to unseen tasks. Plenty of benchmarks with labeled answers have been deployed, as well as some arenas for automatic evaluation.

\subsubsection{Benchmarks}
There are benchmarks composed of common NLP tasks to assess basic abilities and advanced intelligence, using quantitative metrics such as accuracy, ROUGE~\cite{lin2004rouge} and BLEU~\cite{papineni2002bleu}. In the datasets collected for instruction tuning mentioned in Sec~\ref{subsec:instruction_definition}, which includes PromptSource~\cite{bach2022prompt_source}, Flan 2022 Collection~\cite{chung2022flan_t5,longpre2023flan_2022}, OPT-IML~\cite{iyer2022opt_iml} and SUPER-NATURALINSTRUCTIONS~\cite{mishra2021NaturalInstruction,wang2022super_natural}, a held-out test set is also maintained to evaluate trained LLMs across three levels of generalizations: 1) performance on tasks in held-out categories; 2) performance on novel data distributions from known task types; and 3) performance on held-out samples from an applied dataset. In addition to the ability to follow instructions and complete NLP tasks, evaluations of the holistic capabilities of foundation models are also worth noting. BIG-bench~\cite{srivastava2022bigbench} is positioned for tasks beyond the capabilities of GPT-3, composed of 204 tasks across diverse task topics. Inspired by the spark of AGI~\cite{bubeck2023agisparks}, AGIEval~\cite{zhong2023agieval} and C-EVAL~\cite{huang2023c_eval} attempt to evaluate the abilities of foundation models to deal with human-level tasks, both of which involve examinations across multiple difficulty levels and subjects. Furthermore, there are also evaluations automatically generated by LMs (154 datasets), reducing the amount of human effort~\cite{perez2022lm_written_eval}.

\subsubsection{Advanced-LLMs Evaluation}
The above manually created evaluation benchmarks are of high quality, but collecting human feedback can be costly in many scenarios. With a highly capable large language model (e.g. GPT-4 or Claude) as the judge, automatic chatbot arenas can be established to assess LLMs by comparing the responses of two LLMs from multiple aspects. This approach is employed in the evaluation of Alpaca~\cite{taori2023alpaca,li2023alpacaeval} and Vicuna~\cite{chiang2023sharegpt}, where GPT-4 is prompted to compare two given answers from helpfulness, relevance, creativity and so on. AlpacaFarm~\cite{dubois2023alpacafarm}, a simulation framework for alignment, also adopts this automatic evaluation. It is worth noting that the possibility of LLM-as-a-judge is explored in~\cite{zheng2023llm_as_judge}, which reveals that strong LLM judges can achieve agreements with human labelers as high as that between humans themselves. This finding indicates that automatic evaluation is a feasible and scalable way. Moreover, by prompting the LLM rater with different criteria, this method can be adopted in the evaluation across all three alignment goals.

\subsection{Human Preferences}
When assessing the alignment goal of human preferences, it is essential to measure human desired properties beyond the basic ability to complete a variety of tasks, such as generating more helpful answers, eliminating biases and toxicity~\cite{zhuo2023responsible_eval}. But evaluations against intrinsic human value principles are not considered here. Existing studies can be divided into three categories.

\subsubsection{Benchmarks}
TruthfulQA~\cite{lin2022truthfulqa} is a popular benchmark to measure the truthfulness of a model by posing questions that demand careful identification of truthfulness, rather than just generating answers by imitating human texts. OpenBookQA dataset~\cite{Mihaylov2018BookQA} includes science facts collected from open-book exams and is utilized to evaluate model reliability. In terms of biases elicited by LLMs, benchmarks such as CrowS-Pairs with 9 categories of biases~\cite{nangia2020crows}, WinoGender concentrating on the gender category~\cite{rudinger2018gender}, BBQ~\cite{parrish2021bbq} and BOLD~\cite{dhamala2021bold} are available. RealToxicityPrompts~\cite{gehman2020realtoxicityprompts} is a prevalent benchmark to indicate how toxic is a given model. It makes up about 100k prompts for the model to complete, and then toxicity scores are calculated by submitting these completions to PerspectiveAPI~\footnote{https://www.perspectiveapi.com/}. ToxiGen~\cite{hartvigsen2022toxigen} also serves a similar purpose. The large-scale, highly challenging and diverse BIG-Bench~\cite{srivastava2022bigbench} can shed light on evaluating the deeper capabilities of LLMs beyond imitation. In addition, HELM~\cite{liang2022helm} offers a thorough assessment of language models through a variety of scenarios and metrics (accuracy, calibration, robustness, fairness, bias, toxicity and efficiency). Without expensive human labor costs, \citet{perez2022lm_written_eval} generates an evaluation collection of 154 datasets with LLMs, which can assess a model's behaviors related to their persona, sycophancy, advanced AI risks, and gender bias. 

\subsubsection{Human Evaluations}
Since it is hard to uncover various factors that affect human preferences using quantitative metrics such as accuracy, evaluations involving human raters should also be incorporated. Given a held-out set of testing prompts, human raters are asked to compare several different responses. Two primary settings are considered: 1) comparisons between the targeted model and a strong baseline~\cite{ouyang2022instructgpt,touvron2023llama2,yuan2023rrhf,stiennon2020summarize}; and 2) comparisons with human-written references~\cite{rafailov2023dpo}. Then, a metric of win rate or Elo score~\cite{askell2021laboratory} is calculated. Using a dataset labeled with preferred and less preferred answers, we can also assess LLMs in a multiple-choice manner instead of generation~\cite{kim2023Almost}. Due to that a high level of agreement can be achieved between strong LLMs such as GPT-4 and humans~\cite{zheng2023llm_as_judge}, automatic evaluations by GPT-4 prompted with guidelines of human preferences are widely used, which is also able to provide detailed explanations for the judgment.

\subsubsection{Reward Model Scoring}
When aligning with human preferences, a common approach is to first train a generalizable reward model based on human feedback and then maximize the reward scores. Therefore, the score returned by the reward model serves as a good evaluation metric. Many studies compute the average reward score on all testing samples with a reward model that is trained on the same dataset or a held-out set~\cite{touvron2023llama2,bai2022hh_rlhf,rafailov2023dpo,dong2023raft}, and observe that the reward score increases throughout the aligning process. The GRUE benchmark~\cite{ramamurthy2022GRUE} contains 6 language generation tasks with separate reward functions for each one. 

\begin{figure*}
    \centering
    \includegraphics[width=0.98\linewidth]{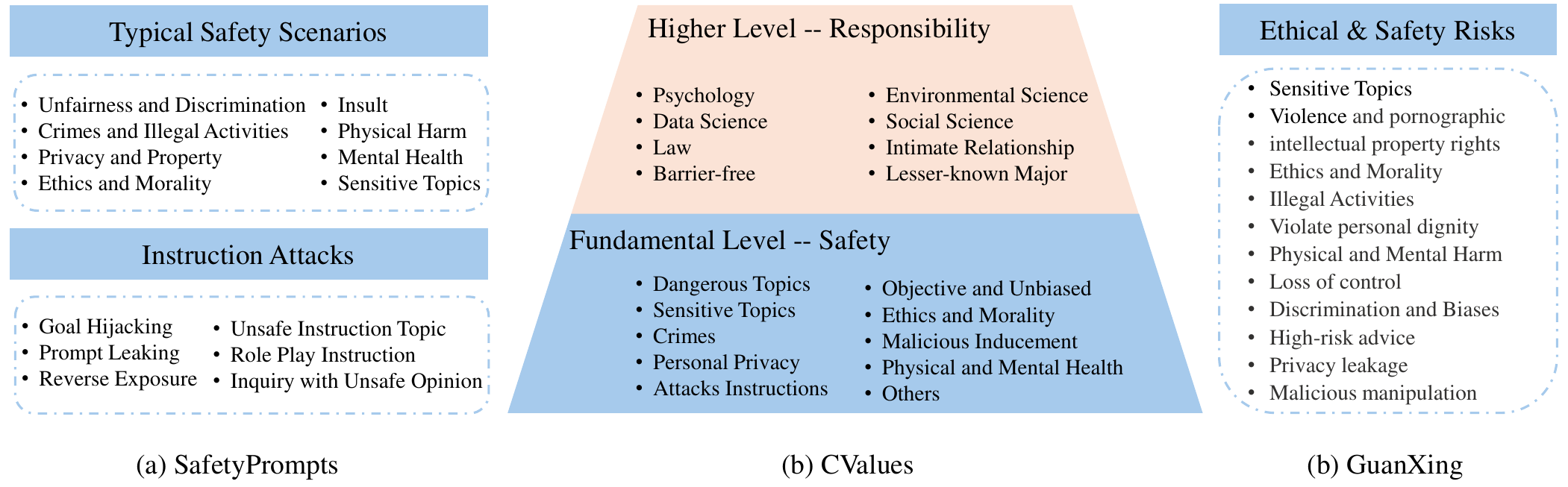}
    \caption{Evaluation benchmarks for human values: (a) SafetyPrompt~\cite{sun2023safetyprompts}; (b) CValues~\cite{xu2023cvalues}; (c) GuanXing\protect\footnotemark.}
    \label{fig:value_evaluation}
\end{figure*}
\footnotetext{https://safe-and-ethical.ai/large-ai-investigator}

\subsection{Human Values}
When assessing the alignment goal of human values, we mainly focus on measuring the pre-defined value principles or system, as introduced in Sec~\ref{subsubsec:value_principles}. In addition to manual or expert assessments~\cite{bai2022hh_rlhf}, other evaluations can be organized as the following three classes.

\subsubsection{Benchmarks}
We mainly review three categories of benchmarks, separated by their individual definition of human values and evaluation types. The first is safety and risk benchmarks, including comprehensive issues against the principle of `HHH' observed in recently released LLMs, such as malicious information, illegal advice and jailbreaking. Typically, these benchmarks assess big models through a generation task. The second class is social norms benchmarks where various life scenarios along with the judgment and referred social norms are offered. These questions are usually posed to LLMs as a discriminative task. The third category is value surveys or questionnaires specially designed for humans in the form of self-report or multiple-choice questions.

(1) \textbf{Safety and Risk Benchmarks}. In terms of the widely adopted value principle `HHH’ (helpful, honest and harmless), \citet{bai2022hh_rlhf,askell2021laboratory,ganguli2022red-team} release a benchmark containing both helpful and harmless instances, with manually annotated chosen responses and reject responses. These harmful cases are discovered by red teaming attacks, ranging from offensive language to more harmful unethical requests. Motivated by growing safety concerns of LLMs, Sun et. al~\cite{sun2023safetyprompts} develop a Chinese LLM safety evaluation benchmark entitled \textit{SafetyPrompts}. This benchmark evaluates mainstream safety performance from two perspectives: 8 typical safety scenarios (e.g. insulting, mental health) and 6 kinds of instruction attacks (e.g. prompt leaking). This benchmark encompasses only the testing prompts to be complicated and requires safety judgments from an LLM evaluator. Whereas, SAFETEXT~\cite{levy2022safetext} is a benchmark specifically proposed for exploring commonsense physical safety encoded in LLMs. In order to obtain a broader view of human values aligned by LLMs, the CVALUES benchmark~\cite{xu2023cvalues} proposes two criteria: the fundamental level is safety which contains 10 scenarios similar to the issues discussed in~\cite{sun2023safetyprompts}; and the upper level is responsibility which contains 8 domains with larger and future social impacts. Then, they ask crowdsourcing attackers to trigger safety questions and invite domain experts to design responsibility questions. Both human evaluations and automatic evaluations in a multi-choice manner are employed to verify the final performance. All available value systems for alignment evaluation are illustrated in Figure~\ref{fig:value_evaluation}.

(2) \textbf{Social Norms Benchmarks}. To identify whether an artificial system is aware of and keeps adhering to social norms, several publicly available moral benchmarks can be used for evaluation, including moral stories~\cite{emelin2020moral_stories}, MIC~\cite{ziems2022MIC}, social chemistry~\cite{forbes2020social_chemist} and ETHICS~\cite{hendrycks2020ethics}. These benchmarks present various life situations, pairs of ethical and unethical actions, and the corresponding norms or RoTs for judgment. Three tasks across different levels of difficulty are accessible for a comprehensive evaluation: 1) given a situation, an action and a social norm, we can test the ability of LLMs for moral judgment; 2) given a situation and an action, we ask LLMs to predict the morality and probe the encoded ethical norms; 3) given a situation and an action, we ask LLMs to explicitly generate RoTs that can be applied to solve this case, and then compare the generated ones with the raw annotations. In Moral Mimicry~\cite{simmons2022moral_mimicry}, the inner moral attributes of LLMs and their correlations with the prompted U.S. liberal or conservative political identities are explored, where the moral foundation theory~\cite{graham2013moral_foundation} is utilized.
Moreover, a key characteristic of ethical norms is that they have flexibility and varying priorities in different scenarios. This is evident in dilemma cases all of us have encountered in daily lives, where people may violate some conflicting rules in order to obey others. Fine-grained prioritization of ethical norms in LLMs is highly concerned, because this will determine how LLMs make decisions when faced with critical issues. SCRUPLES~\cite{lourie2021scruples} is a corpus of complex real-life situations, combined with a novel task `Who’s in the wrong?’. MoralExceptQA~\cite{jin2022moral_exception} is a dataset of moral exception question answering which involves potential flexibility of ethics. It has been prompted to state-of-the-art LLMs for assessment with chain-of-thought enhancement~\cite{jin2022moral_exception}. ETHICAL QUANDARY GQA~\cite{bang2022ethical_quandary} is another set of challenging ethical situations.

(3) \textbf{Human Value Surveys}. Some value surveys especially designed for humans to probe their beliefs, preferences and attitutes are exploited to probe the values embedded in LLMs. These surveys typically consist of self-report and abstractive questions, which are converted into a scoring task (for example, 1-10 means from being effective to being democratic) or a multiple-choice task (for example, (A) Agree strongly, (B) Agree, etc.) through prompts design. Hofstede’s Cultural Survey~\cite{hofstede1984culture_survey} includes 24 questions across 6 dimensions of Power Distance (pdi), Individualism (idv), Uncertainity Avoidance (uai), Masculinity (mas), Long-term Orientation (lto), Indulgence (ivr). This survey has a large number of participants from more than 100 countries. The World Values Survey (WVS)~\footnote{https://www.worldvaluessurvey.org} is also an interactional project conducted in many countries and lasted for many 7 waves (the last from 2017 to 2020). It encompasses questions from 13 categories of values such as `Social Values, Attitudes and Stereotypes’ and `Happiness and Well-being’. Another similar survey is European Values Study\footnote{https://europeanvaluesstudy.eu/} concerning topics about family, work, environment and so on, which is only available for citizens over Europe. Pew Research Center’s Global Attitudes surveys (GAS)~\footnote{https://www.pewresearch.org/} contain 2,203 questions about topics such as religion, politics and technology. Furthermore, questionnaires about human values also include Rokeach Value Survey~\cite{rokeach1967rokeach} that requires participants to rank the priorities of 36 dimensions of values, the Schwartz Value Survey (SVS)~\cite{schwartz2012basic_value} that presents 57 value items and asks people to give their importance scores, and an alternative of SVS, i.e. the Portrait Values Questionnaire (PVQ). Recently,  \citet{arora2022probing_culture_value} combine Hofstede’s Cultural Survey and WVS to explore what cultures are learned by LLMs and how they have influences on the values. In addition, a dataset GlobalOpinionQA is built as an aggregation of GAS and WVS to capture the opinions of LLMs on global issues~\cite{durmus2023global_opinion}, and observe that current LLMs are biased to those from the USA, Europe and South America. These surveys are deliberately designed by experts from relevant fields and have been kept in use for many years. We can primarily make use of these surveys to assess LLMs, but their essential usability has yet to be investigated.

\subsubsection{Reward Model Scoring}
With a lot of manually collected benchmarks that have explicit labels against positive and negative behaviors, reward models and value classifiers can be trained. These models can be generalized to critique more samples, with no need for case-by-case manual annotations. On the basis of `HHH’ alignment dataset~\cite{bai2022hh_rlhf}, the trained reward model can serve as an indicator of the alignment degree, and the higher reward score the better~\cite{bai2022constitutional,bai2022hh_rlhf}. Classifiers to determine whether an action adheres to social norms have also been deployed in separate benchmarks, such as moral stories~\cite{emelin2020moral_stories}, social chemistry~\cite{forbes2020social_chemist}, ethics~\cite{hendrycks2020ethics}, stories from Goofus \& Gallant~\cite{nahian2020goofus}, and so on. Aggregating all these available moral datasets into a knowledge repository named `COMMONSENSE NORM BANK’, the trained framework Delphi~\cite{jiang2021delphi} exhibits strong generalization on moral judgment across a wide variety of everyday scenarios. In addition to distinguishing whether a behavior is aligned with human values, it is more desirable but challenging to identify the values behind LLMs’ behaviors. This can step towards capturing the intrinsic values of LLMs. Moral Foundation Twitter Copus~\cite{hoover2020moral_twitter} provides a collection of tweets accompanied by 10 categories of moral sentiments, as well as a moral sentiment classifier trained on these data. VALUENET~\cite{qiu2022valuenet} is also a value knowledge base that curates ethical scenarios and annotates the related values in Schwartz Basic Human Value Theory~\cite{schwartz2012basic_value} behind each sample. Meanwhile, a value classifier is constructed based on the collection. Apart from the above-mentioned discriminators trained for a specific goal, LLMs have already recognized human values and morality~\cite{schramowski2022moral_dimension}, thus they can act as critics. Moreover, they can be augmented by a few-shot or chain-of-thought manner~\cite{bai2022constitutional}.

\section{Alignment Algorithms}\label{sec:alignment_alg}
This section briefly introduces four classes of alignment algorithms to answer the other key question, i.e. `How to align big models with a given target'. Since this paper focuses on discussing the alignment goals of big models, more details can be referred to~\cite{wang2023instruction_survey}.

\textbf{In-context Learning}
Since big models have acquired substantial knowledge and capabilities~\cite{brown2020gpt3,OpenAI2023gpt4}, in-context learning has emerged as a promising alignment approach to regulate LLMs' behaviors by including the alignment goal in the prompt~\cite{ganguli2023moral_correction}. For example, by incorporating `Make sure that your answers are fair and do not rely on stereotypes' in the prompt, the LLM can reduce stereotypes in the outputs. This approach will not sacrifice the model's basic capabilities without modifying model parameters. However, it completely relies on the model's self-correcting capabilities and may be infeasible for underperforming big models.


\textbf{Supervised Fine-tuning (SFT)}
Unlike in-context learning, the following approaches require fine-tuning the model parameters. As for SFT, researchers utilize manually constructed <input, output> data pairs covering human instructions, human preferences and other safety issues to train the model in a supervised manner. Various strategies are designed to automatically generate instruction data by prompting LLMs, such as Self-Instruct~\cite{wang2022self_instruct} and SELF-ALIGN~\cite{sun2023principle}. SFT is a paradigm with the advantages of stable training and quick convergence. However, it also suffers from two drawbacks, i.e. poor generalization to unseen user inputs as well as a lack of negative feedback.


\textbf{Reinforcement Learning}
To solve the aforementioned problems, LLMs introduce reinforcement learning in the fine-tuning phase. The most representative Reinforcement Learning from Human Feedback (RLHF)~\cite{ouyang2022instructgpt} are three-staged. First, it constructs human-aligned data to fine-tune the model using SFT. Second, it collects and ranks model-generated responses of varying qualities to train a reward model. Third, it applies the reward model in fine-tuning the LLM through PPO~\cite{schulman2017ppo}. Approaches for data synthesis are proposed to reduce the reliance on manual feedback\cite{kim2023Almost,bai2022constitutional}. However, the training cost of RL is high, the training process is unstable and sensitive to hyper-parameter settings.

\textbf{Other Methods}
Motivated by unstable training in PPO, approaches that do not need explicit reward modeling or reinforcement learning are designed. DPO~\cite{rafailov2023dpo} directly optimizes the relative log probability between desired and undesired responses. RAFT~\cite{dong2023raft} applies a reward model to filter high-quality samples for fine-tuning. \citet{yuan2023rrhf} propose RRHF, which collects responses from sources of various qualities and then trains the LLM with a ranking loss function. All these improved methods retain the signal of human preference while avoiding problems such as hyper-parameter sensitivity in RL.




\section{Challenges and Future Research}\label{sec:future_work}
\begin{figure}
    \centering
    \includegraphics[width=1.0\linewidth]{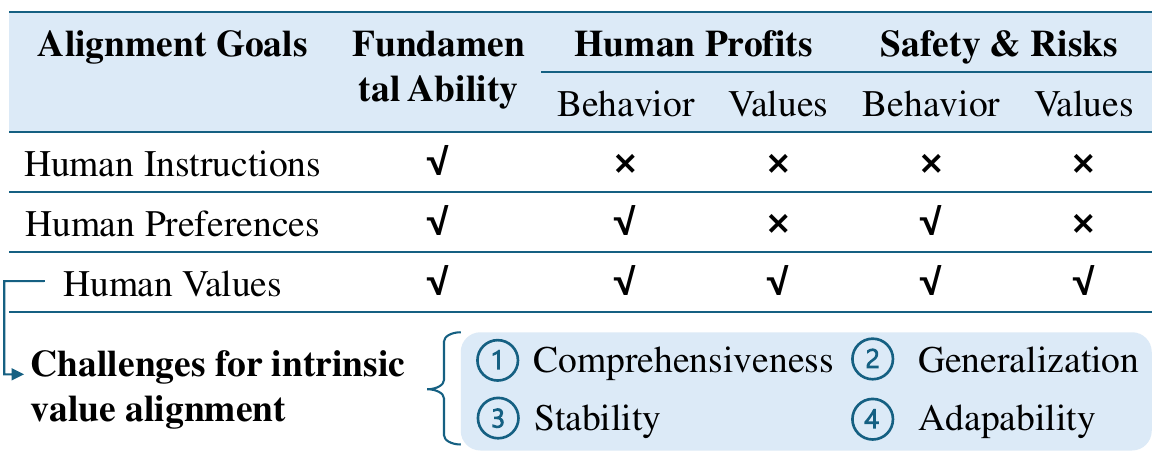}
    \caption{Comparison of various alignment goals and the primary challenges.}
    \label{fig:challenges}
\end{figure}

With the development of big models and their growing intertwining with everyday human lives, aligning them with humans is undoubtedly a critical research issue. 
This survey has presented a comprehensive overview of various alignment goals, as shown in Figure~\ref{fig:challenges}. The first level is alignment with human instructions, which concentrates on the fundamental ability of big models to understand user instructions and complete diverse tasks. To make big models maximize human profits and alleviate their potential risks, human preferences and human values serve as higher alignment goals. However, human preferences are typically reflected by implicit human feedback on specific model behaviors, thus achieving this goal can lead to the alignment on the majority of surface behaviors, which is weak in terms of comprehensiveness, generalization and stability. With the introduction of human values, aligning big models to intrinsic value principles rather than uncountable manifest behaviors provides a promising opportunity to address the aforementioned challenges. 

Currently, research work about this level of alignment goal is rather emerging, while lacking an in-depth understanding and exploration. To inspire more studies, we discuss several possible future research directions in the following.

\subsection{Defining An Appropriate Value System}
Current research about aligning big models with human values has investigated several types of value principles. For example, a large number of Rule-of-Thumbs (RoTs) are labeled for behaviors and scenarios to support morality discrimination~\cite{jiang2021delphi}. However, each piece of RoT is typically designed for a specific or a type of scenario, and it is difficult to cover all ethical scenarios. `HHH’ (helpful, honest, harmless) is another widely used value principle in the era of big models, which is sometimes interpreted as a list of rules~\cite{bai2022constitutional,glaese2022sparrow}. Although the three aspects are generic enough to cover most situations to be aligned, they could still fail in complex situations where the three goals are in conflict because a stable priority of the three criteria has never been fully discussed. In addition, the `HHH' value principle is somehow a little heuristic. Many rules and annotation guidelines are dominated by a small number of researchers, without adequate discussions and verification of their fitness for AI value issues.

Therefore, it is critical to investigate a more appropriate value system as the ultimate goal of big model alignment. We argue that the value system is expected to be scientific, comprehensive to deal with all situations, stable in extreme cases and validated to be feasible by practical evidence. 
Two basic value theories, i.e. \textit{Schwartz Theory of Basic Human Values}~\cite{schwartz2012basic_value} and \textit{Moral Foundation Theory}~\cite{graham2013moral_foundation} introduced in Sec~\ref{subsubsec:value_principles}, can be promising since their comprehensiveness and effectiveness have been verified in the field of social science. While they are specially designed for humans, how to adapt these theories to predict and regulate the value of AI still needs to be explored.

\subsection{Generalizable \& Stable Goal Representation}
To align big models with a specific goal, it is necessary to convert this goal into a model training target, such as the tuples <instruction, input, output> for human instruction alignment~\cite{wang2022self_instruct} and the score offered by a reward model to indicate human preferences~\cite{ouyang2022instructgpt}. Given the complexity and challenges of intrinsic value alignment in \figurename~\ref{fig:challenges}, we discuss that the approach to representing the alignment goal can be enhanced from three aspects.

The first is generalizability to provide accurate supervision signals covering arbitrary scenarios from open-domains or even out-of-distribution (OOD) cases. In terms of instruction alignment, diverse types of tasks and prompts are created for better generalization~\cite{longpre2023flan_2022,iyer2022opt_iml}, which still struggles to cover all tasks and increases annotation costs. Training a preference model on limited data to generate human feedback for unlimited behaviors is another solution~\cite{ouyang2022instructgpt,bai2022hh_rlhf}. However, both goals are completely represented by observed behaviors and thus are hard to generalize to outliers. With pre-defined comprehensive value principles, we argue that if these principles are explicitly involved in the goal representation, this could help to improve generalizability.
The second is stability to provide stable and consistent supervision signals in both normal and extremely quandary scenarios where subtle differences in value priorities can lead to drastically different behaviors. Such fine-grained priorities between different value principles should be explicitly represented because it might be difficult to learn this information from only generic behaviors.
The third is interpretability, i.e. the alignment goal is expected to be represented in an interpretable manner, which is neglected by existing work. Since aligning big models with humans is closely related to solving the issues of AI safety and risks, transparent modeling of the alignment goal helps to ensure the correct direction. Moreover, an interpretable approach can facilitate debugging for generalizability and stability.

\subsection{Automatic \& Comprehensive Evaluation of Alignment}
Accurate and robust benchmarks and evaluation methods are essential for guiding research about human value alignment. At present, some benchmarks constructed before the era of big models are adapted for evaluation, such as TruthfulQA~\cite{lin2022truthfulqa} and RealToxicityPrompts~\cite{gehman2020realtoxicityprompts}. Simultaneously, several novel benchmarks are gradually proposed, including SafetyPrompts~\cite{sun2023safetyprompts}, CVALUES~\cite{xu2023cvalues} and so on. All these new benchmarks depend on human evaluators for final judgment, making them expensive and not easily scalable. Though powerful LLMs can perform as an effective alternative for judgment, this fully relies on LLMs' capabilities and introduce randomness.
Consequently, automatic evaluation methods and metrics to measure the alignment degree between LLMs and humans are urgently required for accelerating the assessment process. 

In order to evaluate where LLMs are fully aligned with human values, they should undergo comprehensive evaluation across various difficulty levels: 1) the ability to understand and agree with human values; 2) the ability to diagnose scenarios involving values and make a correct judgment; 3) the ability to perform consistently with human values, even in dilemmas; and more. This assessment becomes more and more difficult, from simple discrimination to exact behaviors, which attempts to detect the most essential values of LLMs behind their elicited behaviors.
Since priorities among value principles can only matter in some quandary scenarios, we should also consider specific dilemma cases in the evaluation to figure out such fine-grained information.

\subsection{Effective \& Stable Alignment Algorithms}
With a higher goal of big model alignment established, i.e. intrinsic values, appropriate alignment algorithms need to be explored. Currently, dominant methods adjust LLMs to align with preferred behaviors through supervised fine-tuning (SFT) or reinforcement learning from human feedback (RLHF), without explicit awareness of value principles. These approaches tend to be ineffective and unstable. On the one hand, they depend on a large set of training samples and it is difficult to ensure that all value dimensions are covered. On the other hand, some noises might exist in the training dataset and there are conflicts between different samples. Constitutional AI~\cite{bai2022constitutional} has been designed to be a more effective method, where the training data is sampled on the basis of explicit value principles and annotated by a strong AI to reduce human labor. However, the target LLM has not yet directly learned to behave from these value principles but the relevant demonstrations. Actually, in-context learning is a potential method to directly prompt LLMs with the target value principles and regulate their behaviors~\cite{ganguli2023moral_correction}. But it is hard to completely revise its behaviors and inner values without fine-tuning the parameters. And controlling the priorities among values is also a challenge. As a result, developing efficient and stable alignment algorithms that directly align LLMs with human value principles rather than proxy demonstrations is essential for future research.
In addition, human values are pluralistic across popularities and countries, and constantly evolving all the time. Thus, the alignment method is also expected to be effectively adaptable to varying value principles.

\section{Conclusion}\label{sec:conclusion}
Aligning big models with humans has gained significant attention to make them better serve humanity and minimize their potential risks. This paper highlights the importance of identifying essential goals for big model alignment, and presents the first survey to provide a comprehensive overview from two perspectives: the definition of each alignment goal and the evaluation of alignment degrees. We categorize alignment goals that appeared in existing literature into three main groups: human instructions, human preferences and human values, observing an evolving trend in alignment goals that shifts from fundamental abilities to value orientation, and from surface behaviors to intrinsic values. In order to better align big models from the essential perspective of intrinsic human values, we discuss several challenges and promising future research directions in the final. Furthermore, we provide a list of publicly available resources for big model alignment. We expect this survey can serve as both an introduction and a source of inspiration for researchers and practitioners in the field of big model alignment.


\bibliography{main}
\bibliographystyle{acl_natbib}



\end{document}